\pdfoutput=1
\documentclass[11pt]{article}

\usepackage[]{ACL2023}

\usepackage{times}
\usepackage{latexsym}

\usepackage[T1]{fontenc}
\usepackage[utf8]{inputenc}

\usepackage{microtype}

\usepackage{inconsolata}

\usepackage{graphicx}
\usepackage{booktabs}
\usepackage{amsmath}
\usepackage{enumitem}

\title{Massively Multi-Lingual Event Understanding: \\
Extraction, Visualization, and Search}

\author{Chris Jenkins, Shantanu Agarwal, Joel Barry, Steven Fincke, Elizabeth Boschee \\
University of Southern California Information Sciences Institute \\
\{cjenkins, shantanu, joelb, sfincke, boschee\}@isi.edu
}

\begin{document}
\maketitle
\begin{abstract}
In this paper, we present \textsc{ISI-Clear}, a state-of-the-art, cross-lingual, zero-shot event extraction system and accompanying user interface for event visualization \& search. Using only English training data, \textsc{ISI-Clear} makes global events available on-demand, processing user-supplied text in 100 languages ranging from Afrikaans to Yiddish. We provide multiple event-centric views of extracted events, including both a graphical representation and a document-level summary. We also integrate existing cross-lingual search algorithms with event extraction capabilities to provide cross-lingual event-centric search, allowing English-speaking users to search over events automatically extracted from a corpus of non-English documents, using either English natural language queries (e.g. \textit{cholera outbreaks in Iran}) or structured queries (e.g. find all events of type \textit{Disease-Outbreak} with agent \textit{cholera} and location \textit{Iran}). 
\end{abstract}

\section{Introduction}

Understanding global events is critical to understanding the world around us---whether those events consist of pandemics, political unrest, natural disasters, or cyber attacks. The breadth of events of possible interest, the speed at which surrounding socio-political event contexts evolve, and the complexities involved in generating representative annotated data all contribute to this challenge. Events are also  intrinsically global: many downstream use cases for event extraction involve reporting not just in a few major languages but in a much broader context. The languages of interest for even a fixed task may still shift from day to day, e.g.\ when a disease emerges in an unexpected location.

The \textsc{ISI-Clear} (\textsc{Cross-Lingual Event \& Argument Retrieval}) system meets these challenges by building state-of-the-art, language-agnostic event extraction models on top of massively multi-lingual language models. These event models require only English training data (not even bitext---no machine translation required) and can identify events and the relationships between them in at least a hundred different languages. Unlike more typical benchmark tasks explored for zero-shot cross-lingual transfer---e.g. named entity detection or sentence similarity, as in \cite{pmlr-v119-hu20b}---event extraction is a complex, structured task involving a web of relationships between elements in text.

\textsc{ISI-Clear} makes these global events available to users in two complementary ways. First, users can supply their own text in a language of their choice; the system analyzes this text in that native language and provides multiple event-centric views of the data in response. Second, we provide an interface for cross-lingual event-centric search, allowing English-speaking users to search over events automatically extracted from a corpus of non-English documents. This interface allows for both natural language queries (e.g. \textit{statements by Angela Merkel about Ukraine}) or structured queries (\textit{event type = \{Arrest, Protest\}, location = Iraq}), and builds upon our existing cross-lingual search capabilities, demonstrated in \cite{boschee-etal-2019-saral}.

The primary contributions of this effort are three-fold:

\begin{enumerate}[noitemsep,nolistsep]
    \item Strong, language-agnostic models for a complex suite of tasks, deployed in this demo on a hundred different languages and empirically tested on a representative variety of languages.
    \item An event-centric user interface that presents events in intuitive text-based, graphical, or summary forms. 
    \item Novel integration of cross-lingual search capabilities with zero-shot cross-lingual event extraction.
\end{enumerate}

We provide a video demonstrating the \textsc{ISI-Clear} user interface at \url{https://youtu.be/PE367pyuye8}.

% Szósty pakiet sankcji wobec Rosji wszedł w życie, dotyczy przede wszystkim rezygnacji UE z zakupów rosyjskiej ropy.

\begin{figure*}[t]
\centering
\includegraphics[width=12cm]{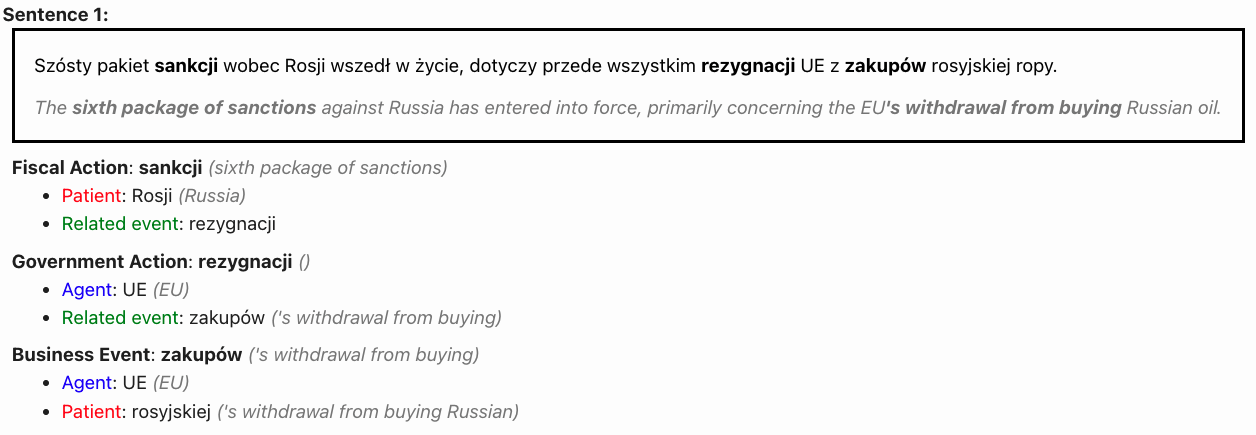}
\caption{
Text-based display of Polish news. 
The user provides only the Polish text. 
To aid an English-speaking user, \textsc{ISI-Clear} displays the extracted event information not only in Polish but also in English.
All processes---including anchor detection, argument extraction, machine translation and span-projection---are carried out in real time.
}
\label{fig:polish_translation}
\end{figure*}

\begin{figure*}[t]
\centering
\includegraphics[width=15cm]{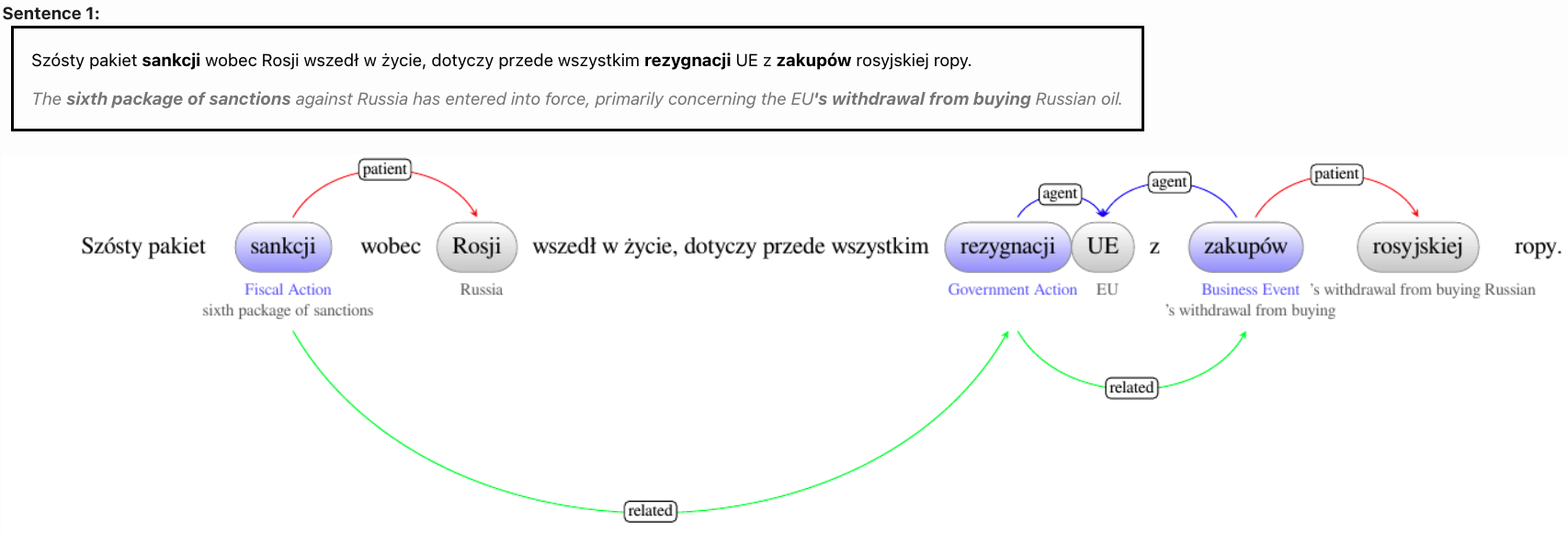}
\caption{
Graph-based display of event information extracted from user provided text in Polish.
}
\label{fig:polish_graphs}
\end{figure*}

\section{User Interface}

\subsection{On-the-Fly Language-Agnostic Event Extraction \& Display}

In our first mode, users are invited to supply their own text in a language of their choice. The system supports any language present in the underlying multi-lingual language model; for this demo we use XLM-RoBERTa \cite{conneau-etal-2020-unsupervised}, which supports 100 languages ranging from Afrikaans to Yiddish.

After submission, the system displays the results in an initial text-based format, showing the events found in each sentence (Figure \ref{fig:polish_translation}). 
For a more intuitive display of the relationships between events, users can select a graphical view (Figure \ref{fig:polish_graphs}). We can easily see from this diagram that the EU is the agent of both the \textit{withdrawal} and the \textit{buying} events, and that the two events are related (the EU is withdrawing from buying Russian oil). 

\begin{figure*}[t]
\centering
\includegraphics[width=14cm]{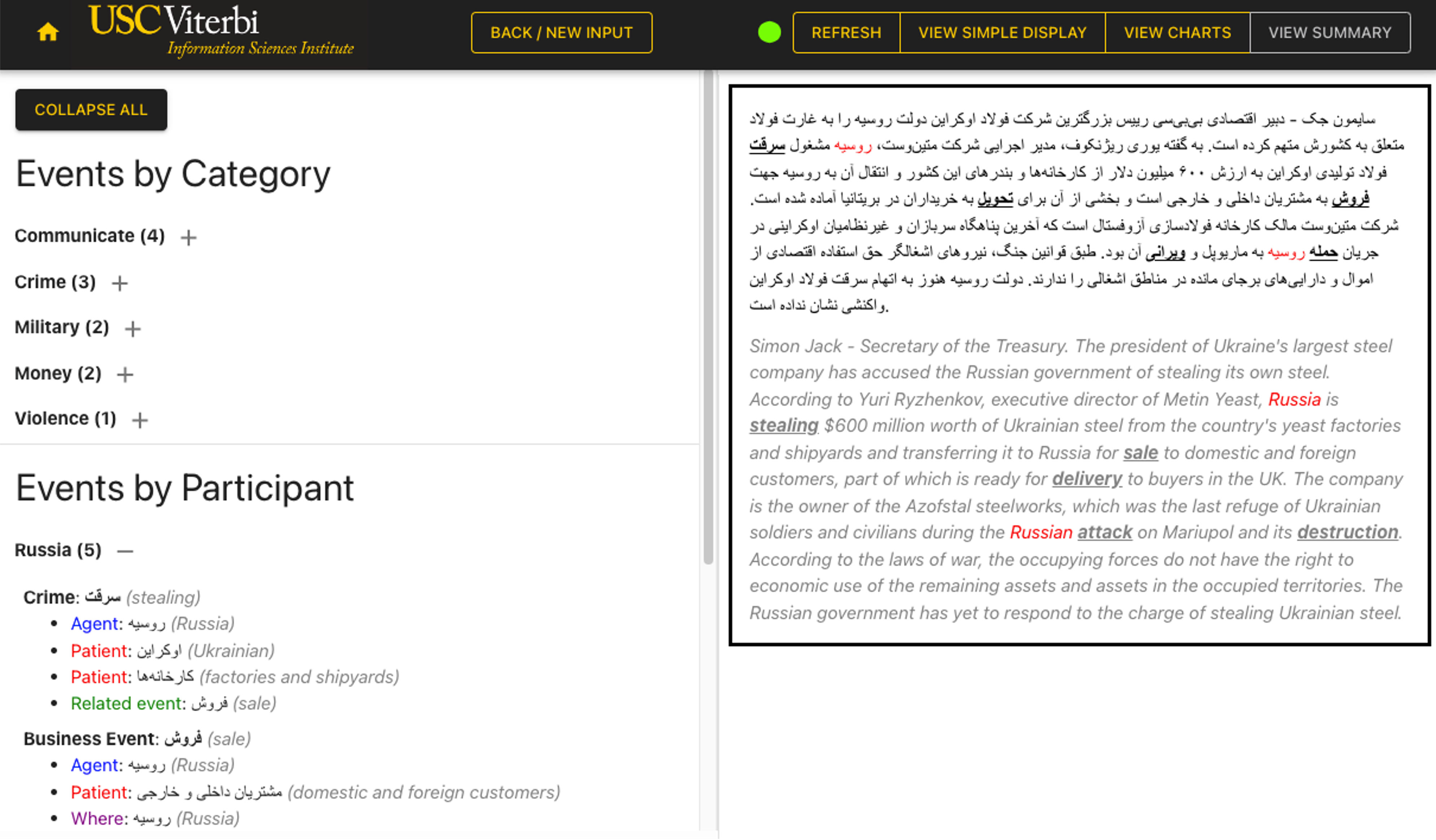} 
\caption{Event-centric summary of Farsi document.}
\label{fig:russian_doc}
\end{figure*}

Finally, the user can see an event-centric summary of the document, choosing to highlight either particular categories of event (e.g., \textit{Crime}, \textit{Military}, \textit{Money}) or particular participants (e.g., \textit{Ukraine}, \textit{Putin}, \textit{Russia}). When one or more categories or participants are selected, the system will highlight the corresponding events in both the original text and, where possible, in the machine translation. An example of a Farsi document is shown in Figure \ref{fig:russian_doc}. Here, the system is highlighting three events in the document where Russia is either an agent or a patient of an event% (specifically: Russia's threat to Western countries, the Russian army targeting Ukraine, and Russian aggression)
. 
For this demo, we use simple heuristics over English translations to group participant names and descriptions; in future work we plan to incorporate a zero-shot implementation of document co-reference to do this in the original language.

\subsection{Cross-Lingual Event-Centric Search}

The second mode of the \textsc{ISI-Clear} demo allows users to employ English queries to search over events extracted from a foreign language corpus. 
%Queries can be expressed either in natural language or constructed using a structured input form, and they operate over an index constructed from system output on a pre-specified set of documents. (We have indexed two separate corpora for this particular demo; see below for more details.)
To enable this, we repurpose our work in cross-lingual document retrieval \cite{barry-etal-2020-searcher} to index and search over event arguments rather than whole documents. A query may specify target \textit{event types} as well as \textit{agent}, \textit{patient}, or \textit{location} arguments; it may also include additional words to constrain the \textit{context}. A sample query might ask for \textit{Communicate} events with the agent \textit{Angela Merkel} and the context \textit{Ukraine}. 

%\textcolor{red}{Shantanu - Do we want to say that the search corpus in this case is pre-indexed. That is do we want to explicitly state that the corpus is not something which the user inputs. It is mentioned in the Corpora sub-section later. But was wondering if it helps the reader to get oriented if it is spelled out up-front.}

\textbf{Query specification.} We allow queries to be specified in two ways. The first simply asks the user to directly specify the query in structured form: using checkboxes to indicate which event types should be included and directly typing in values for each condition (\textit{agent}, \textit{patient}, etc.). A second and more intuitive method allows users to enter a query as natural language. The system processes the query using the \textsc{ISI-Clear} event system and populates a structured query automatically from the results. For instance, if the user enters the phrase \textit{anti-inflation protests in Vietnam}, \textsc{ISI-Clear} will detect a \textit{Protest} event with location \textit{Vietnam} in that phrase. It will turn this result into a query with event type \textit{Protest}, location \textit{Vietnam}, and additional context word \textit{anti-inflation}. 

\textbf{Display.} We display corpus events in ranked order with respect to the user query. The ranking is a combination of system confidence in the underlying extractions (e.g., is this event \textit{really} located in Vietnam?) and system confidence in the cross-lingual alignment (e.g., is \textit{\'etudiants internationaux} really a good match for the query phrase \textit{foreign students}?). To estimate the latter, we rely on our prior work in cross-lingual retrieval, where we developed state-of-the-art methods to estimate the likelihood that foreign text $f$ conveys the same meaning as English text $e$ \cite{barry-etal-2020-searcher}. We note that for locations, we include containing countries (as determined via Wikidata) in the index so that a search for \textit{Iran} will return events happening in, e.g., \textit{Tehran}. More specific details on the ranking functions can be found in Appendix \ref{sec:ranking}.

\begin{figure*}[t]
\centering
\includegraphics[width=16cm]{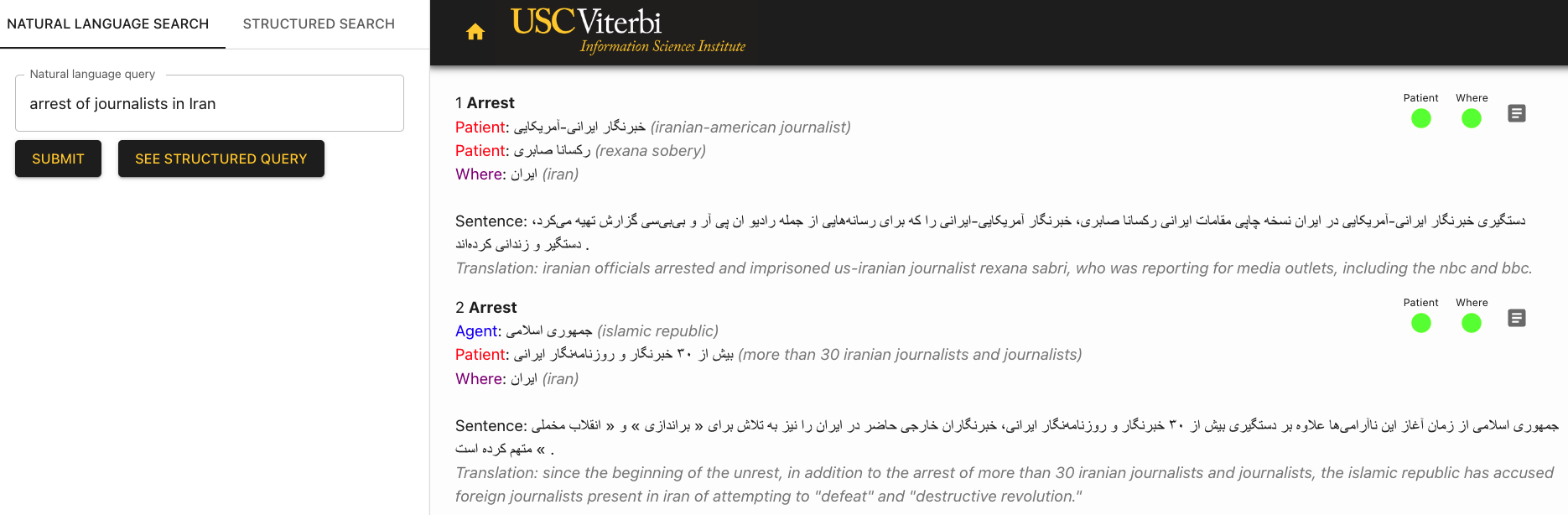}
\caption{Example of search results.}
\label{fig:search}
\end{figure*}

As part of our display, we break down system confidence by query condition---that is, we separately estimate the system's confidence in the \textit{agent} vs., say, the \textit{location}. For each condition, we display a ``traffic light'' indicator that shows the system's confidence in that condition for an event. Red, yellow, and green indicate increasing levels of confidence; black indicates that there is no evidence for a match on this condition, but that other conditions matched strongly enough for the event to be returned. A sample natural language query and search results are shown in Figure \ref{fig:search}.

\textbf{Corpora.} For this demo, we support two corpora: (1) 20,000 Farsi news documents drawn from Common Crawl\footnote{\url{https://commoncrawl.org/}} and (2) $\sim$55K Weibo messages (in Chinese) on the topic of the Russo-Ukrainian crisis \cite{https://doi.org/10.48550/arxiv.2203.05967}. 

\section{Ontology \& Training Data}\label{sec:ontology}

The \textsc{ISI-Clear} demo system is compatible with any event ontology that identifies a set of event types and argument roles. The system expects sentence-level English training data that identifies, for each event, one or more anchor spans and zero or more argument spans (with roles). 

For this demonstration, we use the ``basic event'' ontology and data developed for the IARPA BETTER program (available at \url{https://ir.nist.gov/better/}). The ontology consists of 93 event types and a small set of argument roles (\textit{agent}, \textit{patient}, and \textit{related-event}). In other settings, we have trained and tested the underlying system on the publicly available ACE event ontology\footnote{https://www.ldc.upenn.edu/collaborations/past-projects/ace}, showing state-of-the-art zero-shot cross-lingual results in \cite{fincke2022}. We prefer the BETTER ontology for this demo because of its broad topical coverage and its inclusion of event-event relations (in the form of \textit{related-event} arguments).
The \textsc{ISI-Clear} system is also designed to attach general-purpose \textit{when} and \textit{where} arguments to any event, regardless of ontology; see section \ref{whenwhere}.

\section{System Components}

We present here the highlights of our technical approach, which relies on a collection of strong, language-agnostic models to perform all aspects of event extraction and the classification of relationships between events, as well as machine translation and foreign-to-English projection of event output (for display purposes).

\subsection{Ingest \& Tokenization}\label{sec:token}
Consistent with XLM-RoBERTa, we use Sentence Piece \cite{kudo-richardson-2018-sentencepiece} to tokenize text, and at extraction time, our models label each input subword separately. For languages where words are typically surrounded by whitespace, our system then expands spans to the nearest whitespace (or punctuation) to improve overall performance. If the system produces a conflicting sequence of labels for a single word, we apply simple heuristics leveraging label frequency statistics to produce just one label. 
%We use the Python \textit{langdetect} library to determine whether to expand to whitespace.
%When training (on English), specifically for anchor detection, we have found cross-lingual transfer works best when, during training, we predict from only the first subword in each whitespace- or punctuation-delimited word and ignore those that follow.
%Our intuition is that this is more effective because it allows the learning process to focus on the portion of each word that carries the most useful semantic content---in English, typically the first subword. That semantic pattern, however, is not universal for all languages, so we apply this technique only in English, when training. 
%\textcolor{red}{SCF: Given the page limit, should we shorten or eliminate this (relatively) detailed discussion of tokenization, esp. EN train vs. non-EN test?}

\subsection{Anchor Detection}

\textsc{ISI-Clear} performs anchor identification and classification using a simple beginning-inside-outside (BIO) sequence-labeling architecture composed of a single linear classification layer on top of the transformer stack. For more details please see \cite{fincke2022}.

\subsection{Argument Attachment}

For argument attachment, we consider one event anchor $A$ and one role $R$ at a time. We encourage the system to focus on $A$ and $R$ by modifying the input to the language model. For instance, when $A$=\textit{displaced} and $R$=1 (\textit{agent}), the input to the language model will be \textit{displaced ; 1 </s> Floods < displaced > thousands last month}.
This modification encourages the language model to produce representations of tokens like \textit{thousands} 
that are contextualized by the anchor and role being examined.
The argument attachment model concatenates the language model output vector for each input token with an embedding for event type and applies a linear classifier to generate BIO labels. For more details please see \cite{fincke2022}.

\subsection{Event-Event Relations}

\textsc{ISI-Clear} can handle arbitrary event-event relations within a sentence, including the special case of event co-reference (when a given event has two or more anchor spans). 
We consider one event anchor $A_1$ at a time. Again we modify the input to the language model (by marking $A_1$ with special characters on either side) to encourage the model to consider all other anchors in light of $A_1$. We then represent each event anchor in the sentence (including $A_1$ itself) as a single vector, generated by feeding the language model output for its constituent tokens into a bi-LSTM and then concatenating the bi-LSTM's two final states. (This allows us to smoothly handle multi-word anchors.) To identify the relationship between $A_1$ and $A_2$, if any, we then concatenate the representations for $A_1$ and $A_2$ and pass the result to a linear classifier. The final step optimizes over the scores of all such pairwise classifications to label all relations in the sentence.

\subsection{When \& Where}\label{whenwhere}

The ontology used for this demonstration (described in Section \ref{sec:ontology}) does not annotate \textit{when} and \textit{where} arguments. However, these event attributes are critical for downstream utility. We therefore deploy an ontology-agnostic model that can assign dates and locations to events of any type. To do this, we train a question-answering model to answer questions such as \textit{<s> When/Where did the \{anchor\} happen? </s> Context </s>}. We first train the model on the SQUAD2 dataset \cite{2016arXiv160605250R} and then continue training on the event location and time annotations in the English ACE dataset.
%If the \textsc{ISI-Clear} system is run with an ontology that natively annotates these types of arguments, this step can be skipped.
%\textcolor{red}{Shan: I think the last line can be eliminated (like it is ambiguous what is meant by "run"? Does "run" mean train/decode?).}

\subsection{Machine Translation \& Projection}

All event extraction happens in the target language; no machine translation (or bitext) is required. However, for system output to be useful to English speakers, translation is highly beneficial. Here, we rely on the 500-to-1 translation engine developed by our collaborators at ISI \cite{gowda-etal-2021-many}\footnote{Available at \url{http://rtg.isi.edu/many-eng/}.}. Translation happens after event extraction. We have not optimized this deployment of MT for speed, so we display the results without translation first and then (when the small light in the top toolbar turns green, usually after a few seconds), we can refresh the screen to show results with translations added. 

To project anchor and argument spans into machine translation, we require no parallel data for training. Instead, we leverage the fact that the pre-trained XLM-RoBERTa embeddings are well aligned across languages and have been shown to be effective for word alignment tasks \cite{DBLP:journals/corr/abs-2101-08231}. The similarity of a word in a foreign-language sentence to a word in the parallel English sentence is determined by the cosine distance between the embeddings of the two words. We leverage the Itermax algorithm \cite{jalili-sabet-etal-2020-simalign} to find the best phrase matches. Since we avoid making any bespoke language specific decisions, our projection technique is highly scalable and can project from any of the 100 languages on which XLM-RoBERTa was pre-trained on.

% \section{Ontology \& Training Data}\label{sec:ontology}

% The \textsc{ISI-Clear} demo system is compatible with any event ontology that identifies a set of event types and argument roles. The system expects sentence-level English training data that identifies, for each event, one or more anchor spans and zero or more argument spans (with roles). 

% For this demonstration, we use the ``basic event'' ontology and data developed for the IARPA BETTER program (available at \url{https://ir.nist.gov/better/}). The ontology consists of 93 event types and a small set of argument roles (\textit{agent}, \textit{patient}, and \textit{related-event}). In other settings, we have trained and tested the underlying system on the publicly available ACE event ontology\footnote{https://www.ldc.upenn.edu/collaborations/past-projects/ace}, showing state-of-the-art zero-shot cross-lingual results in \cite{fincke2022}. We prefer the BETTER ontology for this demo because of its broad topical coverage and its inclusion of event-event relations (in the form of \textit{related-event} arguments).
% The \textsc{ISI-Clear} system is also designed to attach general-purpose \textit{when} and \textit{where} arguments to any event, regardless of ontology; see section \ref{whenwhere}.

\section{System Evaluation \& Analysis}

\begin{table*}[]
\centering
\begin{tabular}{@{}llllllllllll@{}}
\toprule
Task & \multicolumn{3}{c}{ACE} & \multicolumn{2}{c}{Basic-1} & \multicolumn{2}{c}{Basic-2} & \multicolumn{4}{c}{Abstract}  \\
\cmidrule(lr){2-4}\cmidrule(lr){5-6}\cmidrule(lr){7-8}\cmidrule(lr){9-12}
Language & en & ar & zh & en & ar & en & fa & en & ar & fa & ko \\ \midrule
Anchors      & 71.2 & 58.1 & 49.6 & 64.2 & 52.5 & 64.6 & 54.3 & 87.4 & 78.3 & 72.5 & 78.9 \\
Arguments    & 72.1 & 51.5 & 51.7 & 64.5 & 51.5 & 71.6 & 64.0 & 69.8 & 45.0 & 45.7 & 45.0 \\
Event coreference  &   --   &  --  &   -- & 83.4 & 67.9 & 86.5 & 65.9 &  --  &  --  & --  &  -- \\ \bottomrule
\end{tabular}
\caption{Component-level accuracy by language / task. Dataset statistics are available in Appendix \ref{sec:appendix-datasets}. 
ACE lacks same-sentence event coreference so those figures are omitted. Event coreference is peripheral to the overall Abstract task; we chose to not model it explicitly and exclude it here.
%Neither ACE nor abstract events have same-sentence event coreference so those figures for those are omitted.
}
\label{tab:accuracy}
\end{table*}

% \begin{table*}[]
% \centering
% \begin{tabular}{@{}lllllllll@{}}
% \toprule
%  & \multicolumn{4}{c}{Per 1000 tokens} & \multicolumn{4}{c}{Per 100 sentences} \\
%  \cmidrule(lr){2-5} \cmidrule(lr){6-9}
%  & en & ar & ru & zh & en & ar & ru & zh \\ \midrule
% Ingest      &  & &  &  &  & &  &  \\
% Anchors     &  & &  &  &  & &  &  \\
% Event-event relations &  & &  &  &  & &  &  \\
% Arguments   &  & &  &  &  & &  &  \\
% When/where  &  & &  &  &  & &  &  \\
% Translation        &  & &  &  &  & &  &  \\
% Projection  &  & &  &  &  & &  &  \\ \midrule
% Avg. tokens per sentence  &  & &  &  &  & &  &  \\ \bottomrule
% \end{tabular}
% \caption{Processing speed broken down by system component.}
% \label{tab:speed}
% \end{table*}

We evaluate our system on a variety of languages and ontologies and compare where possible to existing baselines. Following community practice, e.g.\ \citet{Zhang2019JointEA}, we consider an anchor correct if its offsets and event type are correct, and we consider an argument correct if its offsets, event type, and role find a match in the ground truth. For event coreference (same-sentence only), we consider each anchor pair separately to produce an overall F-score.
% ; ACE is not included as it lacks same-sentence event coreference.

Table \ref{tab:accuracy} provides overall scores in several settings where multi-lingual event annotations are available. All models are trained on English data only. For the ACE data, we follow \cite{huang-etal-2022-multilingual-generative}. The BETTER Basic task is described in Section \ref{sec:ontology}; there are two ontologies (Basic-1 and Basic-2) from different phases of the originating program. The BETTER Abstract task is similar to BETTER Basic, but all action-like phrases are annotated as events, with no further event type specified\footnote{Since abstract events lack event types, we also require anchor offsets to match when scoring arguments.}; valid roles are only \textit{agent} and \textit{patient} \cite{mckinnon-better}. More dataset statistics are found in Appendix \ref{sec:appendix-datasets}.

It is difficult to compare system accuracy across languages; a lower score in one language may reflect a real difference in performance across languages---or just that one set of documents is harder than another. Still, we observe the following. First, performance on anchors seems most sensitive to language choice---for instance, we note that Arabic and Chinese anchor performance on ACE differs by almost 10 points. For arguments, however, non-English performance is relatively consistent given a task---but varies more widely between tasks. Second, we note that cross-lingual performance seems best on anchors, where it exceeds 80\% of English performance for all but one condition. In contrast, argument performance varies more widely, with many conditions below 70\% of English (though some as high as 89\%). 

We also compare against existing published baselines where possible. There are relatively few published results on cross-lingual event anchor detection (and none that we could find on the task of cross-lingual event co-reference as defined here). To benchmark performance on anchors, we turn to MINION \cite{pouran-ben-veyseh-etal-2022-minion}, a multi-lingual anchor-only dataset that uses a derivative of the ACE ontology. For a fair comparison, we retrained our model (tuned for use with XLM-RoBERTa large) with XLM-RoBERTa base; we did not adjust any hyperparameters. Table \ref{tab:minion} shows that the \textsc{ISI-Clear} model performs on average 2.7 points better than the reported MINION numbers for cross-lingual settings. We also show the numbers from our actual demo models (trained with XLM-RoBERTa large) for comparison. 
%on most of the cross-lingual settings. The only exception is Polish, but Polish recovers unusually well when using our standard model (trained with XLM-RoBERTa large), so this could simply be an artifact of the fact that we did not actively tune our system to work with the smaller language model.

\setlength{\tabcolsep}{3pt}
\begin{table}[h]
\centering
\begin{tabular}{@{}l|ccc|c@{}}
\toprule
& \multicolumn{3}{c}{base} & large    \\ \midrule
& MINION & \textsc{ISI-Clear} & $\Delta$ & \textsc{ISI-Clear}    \\ \midrule
en & \textbf{79.5} &  78.9 & -0.6 & 78.0\\ \midrule \midrule
es & \textbf{62.8} &  62.3 & -0.5 & 65.3  \\
pt & \textbf{72.8} &  71.1 & -1.7 & 75.0  \\
pl & \textbf{60.1} &  52.6 & -7.5 & 66.4  \\
tr & 47.2 &  \textbf{52.0} & +4.8 & 56.5  \\
hi & 58.2 &  \textbf{72.2} & +14.0 & 72.7  \\
ko & 56.8 &  \textbf{64.1} & +7.3 & 63.5  \\ \midrule
AVG & 59.7 & \textbf{62.4} & +2.7 & 66.6 \\ \bottomrule
\end{tabular}
\caption{Cross-lingual anchor detection (F1) for MINION dataset, training on English only. Average is across all cross-lingual settings.}
\label{tab:minion}
\end{table}
\setlength{\tabcolsep}{6pt}

For argument detection, much more published work exists, and we show in Table \ref{tab:xgear} that \textsc{ISI-Clear} achieves state-of-the-art performance on all ACE datasets, comparing against the previous state-of-the-art as reported in \citet{huang-etal-2022-multilingual-generative}. 

%Please note that we provide details on system processing speed in Appendix \ref{sec:speed}.

\begin{table}[h]
\centering
\begin{tabular}{@{}l|cc@{}}
\toprule
& X-GEAR &  \textsc{ISI-Clear}    \\ \midrule
en & 71.2 & \textbf{72.1}  \\ \midrule \midrule
ar & 44.8 & \textbf{51.5}  \\
zh & 51.5 & \textbf{51.7}   \\ \bottomrule
\end{tabular}
\caption{Cross-lingual argument detection (F1) for ACE over gold anchors, training on English only.}
\label{tab:xgear}
\end{table}

\section{Related Work}

Several recent demos have presented multi-lingual event extraction in some form, but most assume training data in each target language (e.g. \citet{li-etal-2019-multilingual} or \citet{li-etal-2020-gaia}) or translate foreign-language text into English before processing (e.g. \citet{li-etal-2022-covid}). In contrast, the focus of our demo is making events available in languages for which no training data exists. Other demos have shown the potential of zero-shot cross-lingual transfer, but on unrelated tasks, e.g. offensive content filtering \cite{pelicon-etal-2021-zero}. \citet{akbik-etal-2016-multilingual-information} uses annotation projection from English FrameNet to build target-language models for frame prediction; the focus of the demo is then on building effective queries over language-agnostic frame semantics for extraction. 
%The prior work most similar to \textsc{ISI-Clear} is likely \citet{xia-etal-2021-lome}, where the authors produce and display FrameNet frames cross-lingually. However, in contrast to our work, their supporting models are trained on target-language data, and lack the cross-lingual search-by-query capability which is a key aspect of our demo.
Finally, \citet{xia-etal-2021-lome} also produce FrameNet frames cross-lingually (using XLM-RoBERTa), but in contrast to our work, several of their supporting models use target-language data, and they also supply only a simpler user interface and lack the cross-lingual search-by-query capability that is a key aspect of our demo.

\section{Conclusion}

\textsc{ISI-Clear} provides a monolingual English-speaking user with effective access to global events, both on-demand (extracting events from input of a user's choice) or as a set of indexed documents accessible via cross-lingual search. The system provides a variety of visualizations and modes for engaging with system results. We look forward to future work improving the quality of the underlying components and exploring additional capabilities to cross language barriers and expand access to information around the globe.

\section*{Limitations}
Our core approach is limited by the underyling multi-lingual language model it employs. For this demo, we are therefore limited to the 100 languages that make up the XLM-RoBERTa training set. Performance also varies across languages, tracking in part (though not in whole) with the volume of training data available for each language when building the multi-lingual language model. For instance, anecdotally, the performance on Yiddish (34M tokens in the CC-100 corpus used to train XLM-RoBERTa) is inferior to that of Farsi (13259M tokens). We have provided empirical results for eleven languages and five tasks, but it would be ideal to have a broader set of test conditions; unfortunately, annotated datasets for events are much less common than for simpler tasks like named entity recognition.

A second limitation of our system involves compute requirements. We employ multiple separate components for event extraction (e.g., for anchor detection vs. argument attachment), which increases memory/GPU footprint compared to a more unified system. 

Finally, our system assumes an existing ontology and (English) training data set; it would be interesting to explore zero-shot ontology expansion in future work.

\section*{Ethics Statement}
One important note is that our system is designed to extract information about events that are reported in text, with no judgment about their validity. This can lead a user to draw false conclusions. For instance, the system might return many results for a person $X$ as the agent of a \textit{Corruption} event, but this does not necessarily mean that $X$ is actually corrupt. This should be prominently noted in any use case for this demonstration system or the underlying technologies.

\section*{Acknowledgements}
This research is based upon work supported in part by the Office of the Director of National Intelligence (ODNI), Intelligence Advanced Research Projects Activity (IARPA), via Contract No. 2019-19051600007. The views and conclusions contained herein are those of the authors and should not be interpreted as necessarily representing the official policies, either expressed or implied, of ODNI, IARPA, or the U.S. Government. The U.S. Government is authorized to reproduce and distribute reprints for governmental purposes notwithstanding any copyright annotation therein.

\bibliography{acl2023}

\begin{thebibliography}{22}
\expandafter\ifx\csname natexlab\endcsname\relax\def\natexlab#1{#1}\fi

\bibitem[{Akbik et~al.(2016)Akbik, Chiticariu, Danilevsky, Kbrom, Li, and
  Zhu}]{akbik-etal-2016-multilingual-information}
Alan Akbik, Laura Chiticariu, Marina Danilevsky, Yonas Kbrom, Yunyao Li, and
  Huaiyu Zhu. 2016.
\newblock \href {https://aclanthology.org/C16-2056} {Multilingual information
  extraction with {P}olyglot{IE}}.
\newblock In \emph{Proceedings of {COLING} 2016, the 26th International
  Conference on Computational Linguistics: System Demonstrations}, pages
  268--272, Osaka, Japan. The COLING 2016 Organizing Committee.

\bibitem[{Barry et~al.(2020)Barry, Boschee, Freedman, and
  Miller}]{barry-etal-2020-searcher}
Joel Barry, Elizabeth Boschee, Marjorie Freedman, and Scott Miller. 2020.
\newblock \href {https://aclanthology.org/2020.clssts-1.4} {{SEARCHER}: Shared
  embedding architecture for effective retrieval}.
\newblock In \emph{Proceedings of the workshop on Cross-Language Search and
  Summarization of Text and Speech (CLSSTS2020)}, pages 22--25, Marseille,
  France. European Language Resources Association.

\bibitem[{Boschee et~al.(2019)Boschee, Barry, Billa, Freedman, Gowda, Lignos,
  Palen-Michel, Pust, Khonglah, Madikeri, May, and
  Miller}]{boschee-etal-2019-saral}
Elizabeth Boschee, Joel Barry, Jayadev Billa, Marjorie Freedman, Thamme Gowda,
  Constantine Lignos, Chester Palen-Michel, Michael Pust, Banriskhem~Kayang
  Khonglah, Srikanth Madikeri, Jonathan May, and Scott Miller. 2019.
\newblock \href {https://doi.org/10.18653/v1/P19-3004} {{SARAL}: A low-resource
  cross-lingual domain-focused information retrieval system for effective rapid
  document triage}.
\newblock In \emph{Proceedings of the 57th Annual Meeting of the Association
  for Computational Linguistics: System Demonstrations}, pages 19--24,
  Florence, Italy. Association for Computational Linguistics.

\bibitem[{Conneau et~al.(2020)Conneau, Khandelwal, Goyal, Chaudhary, Wenzek,
  Guzm{\'a}n, Grave, Ott, Zettlemoyer, and
  Stoyanov}]{conneau-etal-2020-unsupervised}
Alexis Conneau, Kartikay Khandelwal, Naman Goyal, Vishrav Chaudhary, Guillaume
  Wenzek, Francisco Guzm{\'a}n, Edouard Grave, Myle Ott, Luke Zettlemoyer, and
  Veselin Stoyanov. 2020.
\newblock \href {https://doi.org/10.18653/v1/2020.acl-main.747} {Unsupervised
  cross-lingual representation learning at scale}.
\newblock In \emph{Proceedings of the 58th Annual Meeting of the Association
  for Computational Linguistics}, pages 8440--8451, Online. Association for
  Computational Linguistics.

\bibitem[{Dou and Neubig(2021)}]{DBLP:journals/corr/abs-2101-08231}
Zi{-}Yi Dou and Graham Neubig. 2021.
\newblock \href {http://arxiv.org/abs/2101.08231} {Word alignment by
  fine-tuning embeddings on parallel corpora}.
\newblock \emph{CoRR}, abs/2101.08231.

\bibitem[{Fincke et~al.(2022)Fincke, Agarwal, Miller, and Boschee}]{fincke2022}
Steven Fincke, Shantanu Agarwal, Scott Miller, and Elizabeth Boschee. 2022.
\newblock Language model priming for cross-lingual event extraction.
\newblock In \emph{Proceedings of the AAAI Conference on Artificial
  Intelligence}.

\bibitem[{Fung and Ji(2022)}]{https://doi.org/10.48550/arxiv.2203.05967}
Yi~R. Fung and Heng Ji. 2022.
\newblock \href {https://doi.org/10.48550/ARXIV.2203.05967} {A weibo dataset
  for the 2022 russo-ukrainian crisis}.

\bibitem[{Gowda et~al.(2021)Gowda, Zhang, Mattmann, and
  May}]{gowda-etal-2021-many}
Thamme Gowda, Zhao Zhang, Chris Mattmann, and Jonathan May. 2021.
\newblock \href {https://doi.org/10.18653/v1/2021.acl-demo.37}
  {Many-to-{E}nglish machine translation tools, data, and pretrained models}.
\newblock In \emph{Proceedings of the 59th Annual Meeting of the Association
  for Computational Linguistics and the 11th International Joint Conference on
  Natural Language Processing: System Demonstrations}, pages 306--316, Online.
  Association for Computational Linguistics.

\bibitem[{Hu et~al.(2020)Hu, Ruder, Siddhant, Neubig, Firat, and
  Johnson}]{pmlr-v119-hu20b}
Junjie Hu, Sebastian Ruder, Aditya Siddhant, Graham Neubig, Orhan Firat, and
  Melvin Johnson. 2020.
\newblock \href {https://proceedings.mlr.press/v119/hu20b.html} {{XTREME}: A
  massively multilingual multi-task benchmark for evaluating cross-lingual
  generalisation}.
\newblock In \emph{Proceedings of the 37th International Conference on Machine
  Learning}, volume 119 of \emph{Proceedings of Machine Learning Research},
  pages 4411--4421. PMLR.

\bibitem[{Huang et~al.(2022)Huang, Hsu, Natarajan, Chang, and
  Peng}]{huang-etal-2022-multilingual-generative}
Kuan-Hao Huang, I-Hung Hsu, Prem Natarajan, Kai-Wei Chang, and Nanyun Peng.
  2022.
\newblock \href {https://doi.org/10.18653/v1/2022.acl-long.317} {Multilingual
  generative language models for zero-shot cross-lingual event argument
  extraction}.
\newblock In \emph{Proceedings of the 60th Annual Meeting of the Association
  for Computational Linguistics (Volume 1: Long Papers)}, pages 4633--4646,
  Dublin, Ireland. Association for Computational Linguistics.

\bibitem[{Jalili~Sabet et~al.(2020)Jalili~Sabet, Dufter, Yvon, and
  Sch{\"u}tze}]{jalili-sabet-etal-2020-simalign}
Masoud Jalili~Sabet, Philipp Dufter, Fran{\c{c}}ois Yvon, and Hinrich
  Sch{\"u}tze. 2020.
\newblock \href {https://doi.org/10.18653/v1/2020.findings-emnlp.147}
  {{S}im{A}lign: High quality word alignments without parallel training data
  using static and contextualized embeddings}.
\newblock In \emph{Findings of the Association for Computational Linguistics:
  EMNLP 2020}, pages 1627--1643, Online. Association for Computational
  Linguistics.

\bibitem[{Kudo and Richardson(2018)}]{kudo-richardson-2018-sentencepiece}
Taku Kudo and John Richardson. 2018.
\newblock \href {https://doi.org/10.18653/v1/D18-2012} {{S}entence{P}iece: A
  simple and language independent subword tokenizer and detokenizer for neural
  text processing}.
\newblock In \emph{Proceedings of the 2018 Conference on Empirical Methods in
  Natural Language Processing: System Demonstrations}, pages 66--71, Brussels,
  Belgium. Association for Computational Linguistics.

\bibitem[{Li et~al.(2022)Li, Gangi~Reddy, Wang, Chiang, Lai, Yu, Zhang, and
  Ji}]{li-etal-2022-covid}
Manling Li, Revanth Gangi~Reddy, Ziqi Wang, Yi-shyuan Chiang, Tuan Lai, Pengfei
  Yu, Zixuan Zhang, and Heng Ji. 2022.
\newblock \href {https://doi.org/10.18653/v1/2022.acl-demo.13} {{COVID}-19
  claim radar: A structured claim extraction and tracking system}.
\newblock In \emph{Proceedings of the 60th Annual Meeting of the Association
  for Computational Linguistics: System Demonstrations}, pages 135--144,
  Dublin, Ireland. Association for Computational Linguistics.

\bibitem[{Li et~al.(2019)Li, Lin, Hoover, Whitehead, Voss, Dehghani, and
  Ji}]{li-etal-2019-multilingual}
Manling Li, Ying Lin, Joseph Hoover, Spencer Whitehead, Clare Voss, Morteza
  Dehghani, and Heng Ji. 2019.
\newblock \href {https://doi.org/10.18653/v1/N19-4019} {Multilingual entity,
  relation, event and human value extraction}.
\newblock In \emph{Proceedings of the 2019 Conference of the North {A}merican
  Chapter of the Association for Computational Linguistics (Demonstrations)},
  pages 110--115, Minneapolis, Minnesota. Association for Computational
  Linguistics.

\bibitem[{Li et~al.(2020)Li, Zareian, Lin, Pan, Whitehead, Chen, Wu, Ji, Chang,
  Voss, Napierski, and Freedman}]{li-etal-2020-gaia}
Manling Li, Alireza Zareian, Ying Lin, Xiaoman Pan, Spencer Whitehead, Brian
  Chen, Bo~Wu, Heng Ji, Shih-Fu Chang, Clare Voss, Daniel Napierski, and
  Marjorie Freedman. 2020.
\newblock \href {https://doi.org/10.18653/v1/2020.acl-demos.11} {{GAIA}: A
  fine-grained multimedia knowledge extraction system}.
\newblock In \emph{Proceedings of the 58th Annual Meeting of the Association
  for Computational Linguistics: System Demonstrations}, pages 77--86, Online.
  Association for Computational Linguistics.

\bibitem[{McKinnon and Rubino(2022)}]{mckinnon-better}
Timothy McKinnon and Carl Rubino. 2022.
\newblock The {IARPA} {BETTER} program abstract task four new semantically
  annotated corpora from {IARPA}’s {BETTER} program.
\newblock In \emph{Proceedings of The 14th Language Resources and Evaluation
  Conference}.

\bibitem[{Pelicon et~al.(2021)Pelicon, Shekhar, Martinc, {\v{S}}krlj, Purver,
  and Pollak}]{pelicon-etal-2021-zero}
Andra{\v{z}} Pelicon, Ravi Shekhar, Matej Martinc, Bla{\v{z}} {\v{S}}krlj,
  Matthew Purver, and Senja Pollak. 2021.
\newblock \href {https://aclanthology.org/2021.hackashop-1.5} {Zero-shot
  cross-lingual content filtering: Offensive language and hate speech
  detection}.
\newblock In \emph{Proceedings of the EACL Hackashop on News Media Content
  Analysis and Automated Report Generation}, pages 30--34, Online. Association
  for Computational Linguistics.

\bibitem[{Pouran Ben~Veyseh et~al.(2022)Pouran Ben~Veyseh, Nguyen, Dernoncourt,
  and Nguyen}]{pouran-ben-veyseh-etal-2022-minion}
Amir Pouran Ben~Veyseh, Minh~Van Nguyen, Franck Dernoncourt, and Thien Nguyen.
  2022.
\newblock \href {https://doi.org/10.18653/v1/2022.naacl-main.166} {{MINION}: a
  large-scale and diverse dataset for multilingual event detection}.
\newblock In \emph{Proceedings of the 2022 Conference of the North American
  Chapter of the Association for Computational Linguistics: Human Language
  Technologies}, pages 2286--2299, Seattle, United States. Association for
  Computational Linguistics.

\bibitem[{{Rajpurkar} et~al.(2016){Rajpurkar}, {Zhang}, {Lopyrev}, and
  {Liang}}]{2016arXiv160605250R}
Pranav {Rajpurkar}, Jian {Zhang}, Konstantin {Lopyrev}, and Percy {Liang}.
  2016.
\newblock \href {http://arxiv.org/abs/1606.05250} {{SQuAD: 100,000+ Questions
  for Machine Comprehension of Text}}.
\newblock \emph{arXiv e-prints}, page arXiv:1606.05250.

\bibitem[{Straka(2018)}]{straka-2018-udpipe}
Milan Straka. 2018.
\newblock \href {https://doi.org/10.18653/v1/K18-2020} {{UDP}ipe 2.0 prototype
  at {C}o{NLL} 2018 {UD} shared task}.
\newblock In \emph{Proceedings of the {C}o{NLL} 2018 Shared Task: Multilingual
  Parsing from Raw Text to Universal Dependencies}, pages 197--207, Brussels,
  Belgium. Association for Computational Linguistics.

\bibitem[{Xia et~al.(2021)Xia, Qin, Vashishtha, Chen, Chen, May, Harman,
  Rawlins, White, and Van~Durme}]{xia-etal-2021-lome}
Patrick Xia, Guanghui Qin, Siddharth Vashishtha, Yunmo Chen, Tongfei Chen,
  Chandler May, Craig Harman, Kyle Rawlins, Aaron~Steven White, and Benjamin
  Van~Durme. 2021.
\newblock \href {https://doi.org/10.18653/v1/2021.eacl-demos.19} {{LOME}: Large
  ontology multilingual extraction}.
\newblock In \emph{Proceedings of the 16th Conference of the European Chapter
  of the Association for Computational Linguistics: System Demonstrations},
  pages 149--159, Online. Association for Computational Linguistics.

\bibitem[{Zhang et~al.(2019)Zhang, Ji, and Sil}]{Zhang2019JointEA}
Tongtao Zhang, Heng Ji, and Avirup Sil. 2019.
\newblock Joint entity and event extraction with generative adversarial
  imitation learning.
\newblock \emph{Data Intelligence}, 1:99--120.

\end{thebibliography}
\bibliographystyle{acl_natbib}

\appendix

\section{Appendix}
\label{sec:appendix}

\subsection{Dataset Statistics}\label{sec:appendix-datasets}

We report results for a variety of different tasks in a variety of different languages. We outline the sizes for these diverse datasets in Tables \ref{tab:training_set_sizes} and \ref{tab:test_set_sizes}. The tasks use five different ontologies; we also report the number of event types for each ontology in Table \ref{tab:ontology_sizes}.

%\begin{table}[h]
\begin{table*}[]
\centering
\begin{tabular}{@{}l|rr|rr}
\toprule
& \multicolumn{2}{c}{Train} & \multicolumn{2}{c}{Development} \\
%& \cmidrule{lr}{2-3}\cmidrule{lr}{4-5}
& \# Characters & \# Events   &\# Characters & \# Events \\  \midrule
ACE   & 1,335,035 & 4,202  & 95,241 & 450 \\
Basic-1   & 171,267 & 2,743   & 35,590 & 560 \\
Basic-2   & 419,642 & 5,995  & 87,425 & 1,214 \\
Abstract    & 557,343 & 12,390  & 67,266 & 1,499 \\
MINION  & 4,388,701 & 14,189 &  544,758 & 1,688 \\ \bottomrule
\end{tabular}
\caption{Size of English training and development sets in number of documents and number of events.}
\label{tab:training_set_sizes}
%\end{table}
\end{table*}

\begin{table}[h]
\centering
\begin{tabular}{@{}l|c|rr@{}}
\toprule
 & Lang. & \# Characters & \# Events   \\ \midrule
ACE & en  & 104,609 & 403 \\
 & ar   & 44,003 & 198 \\
 & zh   & 22,452 & 189 \\
 \midrule
Basic-1 & en   & 33,169 & 569 \\
 & ar   & 238,133 & 5,172\\
 \midrule
Basic-2 & en   & 82,296 & 1,139 \\
 & fa   &  639,6951 & 11,559\\
 \midrule
Abstract & en   & 68,863 & 1,527 \\
 & ar   & 189,174 & 5,339 \\
 & fa   & 607,429 & 15,005 \\
 & ko   & 327,811 & 16,704 \\
 \midrule
 MINION & en    & 554,680 & 1,763 \\
 & es    & 161,159 & 603 \\
 & pt    & 73,610 & 200 \\
 & pl    & 197,270 & 1,234 \\
 & tr    & 175,823 & 814 \\
 & hi    & 57,453 & 151 \\
 & ko    & 332,023 & 164 \\
\bottomrule
\end{tabular}
\caption{Size of test sets in number of documents and number of events.}
\label{tab:test_set_sizes}
\end{table}

\begin{table}[h]
\centering
\begin{tabular}{@{}lr@{}}
\toprule
Ontology & \# of Event Types  \\ \midrule
ACE &  33 \\
Basic-1 & 69 \\
Basic-2 & 93 \\
Abstract & 1 \\
MINION & 16 \\
\bottomrule
\end{tabular}
\caption{Number of event types in each ontology.}
\label{tab:ontology_sizes}
\end{table}

\subsection{Speed}\label{sec:speed}

%Since the same models are used for all languages, we expect speed to be comparable across languages. 
%The main difference is average sentence length, since most of our algorithms do not scale linearly with the number of tokens in a sentence. 
Table \ref{tab:speed} presents speed results for six representative languages, calculated as number of seconds per 100 ``words''. For this exercise we consider words to be the output of UDPipe's language-specific tokenization \cite{straka-2018-udpipe}. 
The primary driver of speed difference is that, given XLM-RoBERTa's fixed vocabulary, different languages will split into more or fewer subwords on average. For instance, an average Korean word will produce at least half again as many subwords than, say, an average Farsi word; this is presumably why 100 words of Korean takes about 70\% longer to process than 100 words of Farsi. On average, for a standard short news article (200 words), we expect to wait about two seconds for extraction and an additional six or seven seconds for MT and projection. We did not optimize our selection of MT package for speed (e.g., it decodes one sentence at a time instead of batching); this could easily be updated in future work to be more efficient.

\begin{table}[h]
\centering
\begin{tabular}{@{}l|llllll@{}}
\toprule
 & en & ar & fa & ko & ru & zh   \\ \midrule
Event   & 1.1 & 1.0 & 0.9 & 1.5 & 0.8 & 1.1    \\
Display & n/a & 2.6 & 2.8 & 4.1 & 3.4 & 3.9  \\\bottomrule
%Event   & 10.8 & 10.1 & 8.9 & 15.3 & 8.4 & 10.8    \\
%Display & n/a & 25.6 & 27.5 & 40.7 & 33.9 & 39.0  \\\bottomrule
\end{tabular}
\caption{Processing speed (seconds per 100 words). Event processing includes ingest, tokenization, anchors, arguments, event-event relationships, and when/where extraction. Display processing includes components solely required for display (MT and projection). We use 11GB GTX 1080Ti GPUs for extraction/projection and use a 48GB Quadro RTX 8000 GPU for MT.}
\label{tab:speed}
\end{table}

\subsection{Search Ranking}\label{sec:ranking}

\textsc{ISI-Clear} extracts a large number of events from the documents indexed from search, some of which vary in quality and some of which will match more or less confidently to an English query. The ranking function described here significantly improves the usability of our search results.

The goal of our search ranking function is to rank each extracted event $E$ with respect to a user query $Q$. To calculate $score(Q, E)$, we combine two separate dimensions of system confidence:
\begin{enumerate}[noitemsep,nolistsep]
    \item \textit{Cross-lingual alignment confidence (CAC)}: are the components of $E$ reasonable translations of the query terms? For instance, is \textit{\'etudiants internationaux} a good match for the query phrase \textit{foreign students}? Here, we assume the existence of a cross-lingual retrieval method $cac(e, f)$ that estimates the likelihood that foreign text $f$ conveys the same meaning as English text $e$, as in our prior work \cite{barry-etal-2020-searcher}.
    \item \textit{Extraction confidence (EC)}: how likely is it that the elements of $E$ were correctly extracted in the first place? Here we use confidence measures (denoted $ec$) produced by individual system components.
\end{enumerate}

To combine these dimensions, we consider each query condition separately (summing the results). For simplicity we describe the scoring function for the \textit{agent} condition: 
\begin{multline*}
score(Q_{agent}, E_{agent}) = \\
\beta * ec(E_{agent}) * cac(Q_{agent}, E_{agent})\ + \\
(1 - \beta) * cac(Q_{agent}, E_{sentence})
\end{multline*}
The first term of this equation captures the two dimensions described above. The second term allows us to account for agents missed by the system, letting us give ``partial credit'' when the user's search term is at least found in the nearby context (e.g., in $E_{sentence}$). Based on empirical observation, we set $\beta$ to 0.75. 

We follow the same formula for \textit{patient} and \textit{location}. For \textit{context} we use only the final term $cac(Q_{topic}, E_{sentence})$ since \textit{context} does not directly correspond to an event argument. 

For now, event type operates as a filter with no score attached; in future work we will incorporate both the system's confidence in the event type as well as a fuzzy match over nearby event types (e.g., allowing for confusion between \textit{Indict} and \textit{Convict}).

\end{document}